\documentclass[lettersize,journal]{IEEEtran}
\usepackage{amsmath,amsfonts}
\usepackage{algorithmic}
\usepackage{algorithm}
\usepackage{array}
\usepackage[caption=false,font=normalsize,labelfont=sf,textfont=sf]{subfig}
\usepackage{textcomp}
\usepackage{stfloats}
\usepackage{url}
\usepackage{verbatim}
\usepackage{graphicx}
\usepackage{cite}
\usepackage{booktabs}
\usepackage{amssymb}
\usepackage{multirow}

\begin{document}

\title{MsFIN: Multi-scale Feature Interaction \\ Network for Traffic Accident Anticipation}

\author{Tongshuai Wu, Chao Lu*, ~\IEEEmembership{Member,~IEEE}, Ze Song, \\Yunlong Lin, Sizhe Fan, Xuemei Chen*

\thanks{This research is supported by the National Natural Science Foundation of China under Grants 52372405 and 61703041, and is also supported by China Scholarship Council (CSC).}

\thanks{Tongshuai Wu, Chao Lu, Ze Song, Yunlong Lin, Sizhe Fan and Xuemei Chen are with the School of Mechanical Engineering, Beijing Institute of Technology, Beijing 100081, China (Email: wutongshuai84@gmail.com; chaolu@bit.edu.cn; 3120230312@bit.edu.cn; yunlonglin@bit.edu.cn; sizhefan@bit.edu.cn; chenxue781@bit.edu.cn)}

\thanks{(*Corresponding authors: Chao Lu and Xuemei Chen.)}}




\maketitle

\begin{abstract}
With the widespread deployment of dashcams and advancements in computer vision, developing accident prediction models from the dashcam perspective has become critical for proactive safety interventions. However, two key challenges persist: modeling feature-level interactions among traffic participants (often occluded in dashcam views) and capturing complex, asynchronous multi-temporal behavioral cues preceding accidents. To deal with these two challenges, a Multi-scale Feature Interaction Network (MsFIN) is proposed for early-stage accident anticipation from dashcam videos. MsFIN has three layers for multi-scale feature aggregation, temporal feature processing and multi-scale feature post fusion, respectively. For multi-scale feature aggregation, a Multi-scale Module is designed to extract scene representations at short-term, mid-term and long-term temporal scales. Meanwhile, the Transformer architecture is leveraged to facilitate comprehensive feature interactions. Temporal feature processing captures the sequential evolution of scene and object features under causal constraints. In the multi-scale feature post fusion stage, the network fuses scene and object features across multiple temporal scales to generate a comprehensive risk representation. Experiments on DAD and DADA datasets show that MsFIN significantly outperforms state-of-the-art models with single-scale feature extraction in both prediction correctness and earliness. Ablation studies validate the effectiveness of each module in MsFIN, highlighting how the network achieves superior performance through multi-scale feature fusion and contextual interaction modeling.

\end{abstract}


\section{Introduction}
\IEEEPARstart{R}{oad} traffic accidents cause considerable economic losses for individuals, their families, and nations as a whole. According to the World Health Organization, approximately 1.19 million people die each year as a result of road traffic accidents\cite{1}. With the development of sensing technologies, Traffic Accident Anticipation (TAA) has emerged as an effective approach to mitigate the incidence of road traffic accidents \cite{bao2020uncertainty}. TAA utilizes data collected from vehicle-mounted or roadside sensors to perceive traffic scenes and predict accidents at the earliest feasible stage. This capability empowers drivers or autonomous systems to implement preventive measures against traffic accidents.

Compared with other vehicle-mounted sensors such as radar and LiDAR, cameras offer a more cost-effective solution for TAA. As a result, vision-based TAA using data collected from cameras has been one of the most widely used TAA technologies in traffic scenes. This approach prefers the early prediction of accidents from video sequences, thereby enabling extended time windows for implementing proactive safety interventions.

Early vision-based TAA research primarily focused on surveillance perspectives. These methods typically estimated collision risk by predicting the future trajectories of traffic participants and calculating their physical proximity. For example, a 3D model-based vehicle tracking approach for traffic accident prediction was proposed in \cite{hu2003traffic}. The potential collisions risk are identified by analyzing the geometric intersections of vehicle bounding boxes over consecutive time steps. To reduce false alarms caused by purely geometric detection, an adaptive fuzzy neural network was further introduced to learn trajectory patterns \cite{hu2004traffic}. By dynamically adjusting neuron weights through fuzzy membership functions, the method effectively addresses the issue of local oscillations in self-organizing networks. Similarly, some methods \cite{shan2017vehicle,haris2021vehicle,chavan2021collide} predicted participant trajectories and evaluate potential spatial conflicts to identify scene risk. While effective under surveillance perspectives, these methods struggle to generalize to ego-centric scenes, where trajectory prediction is often hindered by occlusions and limited observability.

\IEEEpubidadjcol

In recent years, with the rapid advancement of autonomous driving technologies, vision-based TAA from dashcam perspectives has gradually emerged as a prominent research focus. Unlike fixed-view surveillance footage, dashcam videos frequently exhibit occlusions among road participants, making physical distance-based measurements less reliable. To address this challenge, recent studies have shifted focus toward modeling the interactions between traffic participants\cite{fang2023vision}. Graph-based networks have been widely adopted for modeling interactions among traffic participants \cite{li2022driver, yu2021scene, song2024dynamic, thakur2024graph, malawade2022spatiotemporal, lin2023continual}, they often suffer from limited flexibility. The predefined graph topology constrains the interaction patterns that the model can capture. Some studies have captured causal factors by distributing soft attention to objects in traffic scene\cite{karim2022dynamic,chan2017anticipating,fatima2021global,zeng2017agent}. DSA-RNN introduces a dynamic spatial attention mechanism that assigns soft attention weights to candidate objects in each frame\cite{chan2017anticipating}. Building upon this, DSTA integrates the dynamic spatial attention and dynamic temporal attention\cite{karim2022dynamic}. These methods mainly reflect the importance of individual participants to the scene through dynamic weighting, but fail to consider the connections among participants. This makes it difficult for the model to capture multi-object collaborative behaviors or potential conflicts.

Traffic accidents are often caused by a combination of environmental factors and the behaviors of participants. The behavioral relationships among participants implicitly contain underlying information about conflict. Considering the environment in which participants are situated helps the model understand the different risks exhibited by the same behavior in varying contexts. Therefore, modeling the feature interactions among participants as well as between participants and scene can enhance the risk perception capability of models. 

In addition, existing vision-based TAA rely solely on single-time-scale modeling\cite{karim2022toward,karim2023attention,wang2023gsc}. XAI is proposed in \cite{karim2022toward}. It integrates Gated Recurrent Unit (GRU) \cite{chung2014empirical} and Gradient-weighted Class Activation Mapping (Grad-CAM). GRU is employed to model the temporal evolution of scene information. And its hidden state is updated by mean pooling over the past $M$ frames. Grad-CAM generates saliency maps to interpret the decision of the model, thereby enhancing the interpretability of accident anticipation. AM-NET is proposed for early localization of risk-relevant participants in driving sequence \cite{karim2023attention}. The network feeds object features and optical flow features into two separate GRU. An attention module is used to weight the hidden representations from the past $M$ frames and update the current hidden state. Finally, the combined hidden representation is used to predict the risk score for each object. These methods adopt single temporal window. Traffic accidents typically involve multi-stage dynamic evolution process. For instance, rear-end collision typically progresses through three stages: normal following, lead vehicle deceleration and emergency braking. Relying on a single-time-scale makes it difficult for the model to comprehensively understand and anticipate such evolving risks.

To address the above limitations, a novel vision-based TAA model named Multi-scale Feature Interaction Network (MsFIN) is proposed for TAA. The network leverages the Multi-scale Module to capture scene information at different temporal scales. Meanwhile, the feature interactions among participants and between participants and the traffic scene are modeled using the Transformer architecture \cite{vaswani2017attention}. Experimental results demonstrate that the proposed MsFIN achieves superior performance on two benchmark datasets, DAD\cite{chan2017anticipating} and DADA\cite{fang2021dada}. MsFINoutperforms existing single-scale interaction models in both correctness and earliness of accident anticipation. The contributions of this paper are are as follows:

\begin{itemize}
\item{A Multi-scale Module is proposed, which employs a parallel pooling strategy to capture traffic scene dynamics across short-term, mid-term, and long-term scales. Specifically, short-term scale features help detect sudden high-risk events. Mid-term scale features track the progressive development of risky situations. Long-term scale features mitigate the forgetting of early-stage cues. The multi-scale features exhibit complementarity across different types of accidents, which effectively enhances the earliness of accident anticipation.}

\item{A comprehensive feature interaction framework is constructed. Self-attention models the interactions among participants. Cross-attention captures interactions between participants and multi-scale scene information. These mechanisms effectively enhance the model's capability to represent accident risk.}

\item{An adaptive loss function is proposed by introducing focal loss into the standard exponential loss. The modified loss effectively mitigates the issue of easy-to-classify samples dominating the gradient updates and enhances learning from hard positive examples.}
\end{itemize}

The remainder of this paper is organized as follows: Section II details the architecture and components of proposed MsFIN. Section III presents the experimental setup, implementation details, and a thorough evaluation of the proposed method. Finally, Section IV concludes the paper and discusses future work.

\section{Proposed Method}
This section begins with the problem formulation of the TAA task. A detailed introduction is provided to the core components of the proposed MsFIN network, including multi-scale feature aggregation, temporal feature processing and multi-scale feature post-fusion. The loss functions employed during training are also described.

\subsection{Problem Formulation}

\begin{figure*}[t]
\centering
\includegraphics{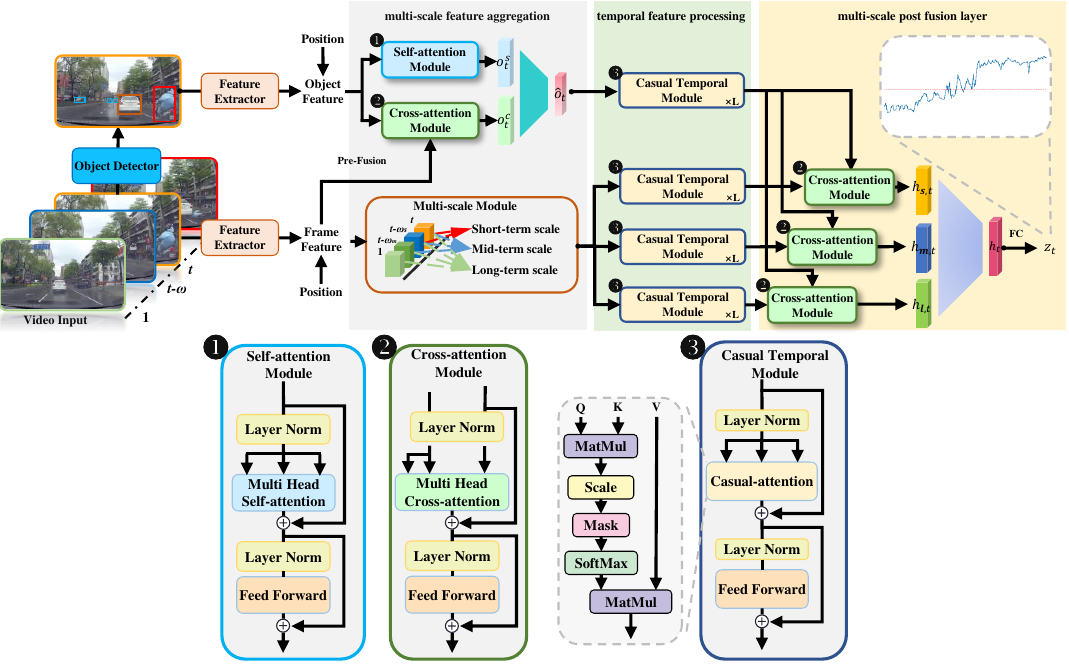}
\caption{Overview of the Multi-scale Feature Interaction Network.}
\label{MsFIN_framework}
\end{figure*}

MsFIN primarily consists of multi-scale feature aggregation, temporal feature processing, and multi-scale feature post-fusion, as illustrated in Fig.~\ref{MsFIN_framework}. These components are designed to capture rich risk representations of traffic scenes from different temporal scales. Consistent with existing vision-based TAA methods, this paper formulates the Vision-TAA as a classification problem. Specifically, given a driving video sequence with $T$ frames:

\begin{equation}
\mathcal{V} = \{I_1, I_2, \dots, I_T\}
\end{equation}

\noindent where $I_t$ denotes the image at time step $t\in \left[ 1,T \right]$. Each frame undergoes object detector to extract participants information. The image and participant information are separately processed through a feature extractor. This process results in a frame-level feature sequence $\mathcal{F} \in \mathbb{R}^{T \times d}$ that contain scene information and object-level feature sequence $\mathcal{O} \in \mathbb{R}^{T \times N \times d}$ that represent participant information. To comprehensively characterize accident risk, MsFIN employs a multi-scale feature aggregation layer. Object-level features are processed through the Self-attention Module (SaM) and Cross-attention Module (CaM) to achieve feature interaction among participants and from scene to each participant.

\begin{equation}
\widehat{\mathcal{O}} = \Psi_{\text{agg}}\left( \text{SaM}(\mathcal{O}), \text{CaM}(\mathcal{O}, \mathcal{F}) \right)
\end{equation}

\noindent In parallel, MsFIN simultaneously constructs temporally-aware scene representations by aggregating frame-level features across multiple temporal scales. This allows the model to capture evolving traffic dynamics and temporal cues relevant for accident risk:

\begin{equation}
\widehat{\mathcal{F}}_{p} = \Psi_{p}\left( \left\{ \mathcal{F}_{t^{\prime}} \right\}_{t^{\prime} \in \Omega_{p}(t)} \right)
\end{equation}

\noindent where $p \in \left\{ \text{s}, \text{m}, \text{l} \right\}$ denotes the temporal scales, corresponding to short-term, medium-term, and long-term windows, respectively. $\Omega_{p}(t)$ denotes the temporal window at scale $p$ centered at time $t$. To model temporal dependencies while respecting causality, the aggregated features are fed into temporal feature processing.

\begin{equation}
\begin{bmatrix}
\widetilde{\mathcal{O}} \\ 
\widetilde{\mathcal{F}}_{p}
\end{bmatrix} = \text{CTM}\left(
\begin{bmatrix}
\widehat{\mathcal{O}} \\
\widehat{\mathcal{F}}_{p}
\end{bmatrix}
\right)
\end{equation}

\noindent This layer is implemented using stacked Causal Temporal Module (CTM), ensuring that the output at each frame only depends on current and past information. Finally, to fuse multi-scale temporal cues into a unified representation, a multi-scale feature post-fusion is introduced. It aggregates participant behavior variations across different temporal scales by performing cross-attention with participant features as key and value:

\begin{equation}
\mathcal{H} = \bigoplus\limits_{p \in \{s,m,l\}} \text{CaM}\left(\widetilde{\mathcal{F}}_p, \widetilde{\mathcal{O}}\right) 
\end{equation}

\noindent where $\mathcal{H} \in \mathbb{R}^{T \times D}$ denotes the multi-scale fused representation. This unified representation is then passed through a Multilayer Perceptron (MLP) to obtain a frame-wise accident risk probability sequence:

\begin{equation}
\mathcal{R}=[{{z}_{1}},{{z}_{2}},\dots,{{z}_{T}}]   
\end{equation}

\noindent where $z_t>\tau$ denotes that an accident is considered to have occurred at time step $t$.

\subsection{Multi-scale Feature Aggregation}
For each driving sequence, the same object detection and feature extraction method as in \cite{chan2017anticipating} is adopted. This ensures that the comparison focuses on the accident anticipation modeling algorithm rather than the object detection or feature extraction techniques. The specific feature extraction steps will be detailed in the next section. After feature extraction and the addition of learnable positional encodings, the frame-level feature sequence and the object-level feature sequence:

\begin{equation}
\mathcal{O} = \{o_1, o_2, \dots, o_T\}
\end{equation}

\begin{equation}
\mathcal{F} = \{f_1, f_2, \dots, f_T\}
\end{equation}

\noindent where $f_t \in \mathbb{R}^{d}$ denotes the frame-level feature at time step $t$. And $o_t \in \mathbb{R}^{N \times d}$ represents the object-level features of $N$ detected participants in the same frame. These features are first passed through multi-scale feature aggregation to enhance their representational capacity. The multi-scale feature aggregation is composed of object-level feature aggregation and scene-level temporal aggregation. 

In object-level feature aggregation process $o_t$ are passed through two attention modules. First, a Self-attention Module (SaM) are applied to model the feature interaction among participants. By leveraging a multi-head self-attention mechanism, the SaM enables each participant to dynamically attend to other participants in the same frame. This captures complex spatial dependencies and contextual behaviors. For $o_t$  the attention representation is computed as follows:

\begin{equation}
\label{equ1}
\bar{o}_{t} = \text{LayerNorm}(o_t)
\end{equation}

\begin{equation}
Q = W_Q \bar{o}_t, \quad K = W_K \bar{o}_t, \quad V = W_V \bar{o}_t
\end{equation}

\begin{equation}
\text{Attn}(Q, K, V) = \text{LayerNorm} \left( \frac{QK^\top}{\sqrt{d}} \right) V
\end{equation}

\begin{equation}
o_t^{\text{A}} = \text{LayerNorm} \left( \bar{o}_t + \text{Attn}(Q, K, V) \right)
\end{equation}

\begin{equation}
\label{equ2}
o_{{t}}^{\text{S}} = o_t^{\text{A}} + \text{FeedForward}(o_t^{\text{A}})
\end{equation}

\noindent where $W_Q, W_K, W_V \in \mathbb{R}^{d \times d}$ are learnable projection matrices. $Q, K, V$ are the queries keys and values respectively. In parallel with SaM, Cross-attention Module (CaM) uses object-level features as queries and frame-level features as keys and values. By computing attention weights, the scene information effectively supplements the participants features. The resulting feature sequence are denoted as ${\mathcal{O}^{\text{S}}}$ and ${\mathcal{O}^{\text{C}}}$, respectively. After concatenation, the features are projected through a linear layer to produce the aggregated object-level feature sequence representation $\widehat{\mathcal{O}}\in {{\mathbb{R}}^{T\times N\times d}}$$:$

\begin{equation}
\widehat{\mathcal{O}}=W(\text{concat}({\mathcal{O}^{S}},{\mathcal{O}^{C}}))+B
\end{equation}

\noindent where $W$ and $B\in {{\mathbb{R}}^{d\times d}}$ are learnable parameters of the fully connected layers.

\begin{figure}[t]
\centering
\includegraphics{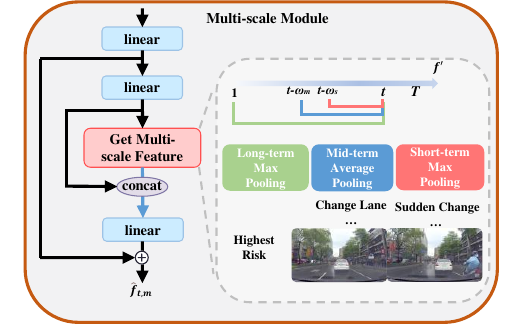}
\caption{The Multi-scale Module.}
\label{MsM}
\end{figure}

Scene-level temporal aggregation is conducted to capture evolving dynamics of the traffic scene. It is achieved by the Multi-scale Module (MsM), which employs a parallel pooling mechanism to extract multi-scale features across different temporal scales. As shown in Fig.~\ref{MsM} three types of temporal context (short-term, mid-term, and long-term) are captured using short-term max pooling, mid-term average pooling, and long-term max pooling, respectively. For $t$ time step frame-level feature ${{f}_{t}}$, the multi-scale features are computed as:

\begin{equation}
f_t' = W f_t + B
\end{equation}

\begin{equation}
f_{t,s} = \max_{t' \in (t - w_s,\, t]} \; f_{t'}'
\end{equation}

\begin{equation}
f_{t,m} = \frac{1}{w_m} \sum_{t' \in (t - w_m,\, t]} f_{t'}'
\end{equation}

\begin{equation}
f_{t,l} = \max_{t' \in (0,\, t]} \; f_{t'}'
\end{equation}

\noindent where $ f_{t,s} $, $ f_{t,m} $, and $ f_{t,l} $ denote the short-term, mid-term, and long-term temporal representations at time step $ t $, respectively. ${{w}_{s}}$ and ${{w}_{m}}$ represent the sizes of the short-term and mid-term scales time windows. Then they are processed through linear layer and residual connection, resulting in frame-level aggregated features at different scale $\widehat{f}_{t,p}$.

\begin{equation}
\widehat{f}_{t,p}=W(\text{concat}({{f}_{t,p}},f_t'))+B+{{f}_{t}}
\end{equation}

\noindent The final aggregated scene feature sequences at short-term, mid-term, and long-term temporal scales are denoted as $ \widehat{\mathcal{F}}_{s} $, $ \widehat{\mathcal{F}}_{m} $, and $ \widehat{\mathcal{F}}_{l} $, respectively:

\begin{equation}
\widehat{\mathcal{F}}_{p} = \{ \widehat{f}_{1,p}, \widehat{f}_{2,p}, \dots, \widehat{f}_{T,p} \}
\end{equation}

\noindent Short-term scale focuses on transient high-risk events, such as sudden braking or abrupt lane changes. Mid-term scale captures moderate changes in risk, including lane merging or gradual acceleration. Long-term scale helps retain awareness of earlier high-risk situations that may influence future outcomes.

\subsection{Temporal Feature Processing}
Transformer architecture has demonstrated outstanding performance in video sequence tasks. The Transformer architecture has been applied to video classification tasks, achieving high performance on multiple mainstream benchmark datasets \cite{arnab2021vivit}. Traditional RNN (e.g., LSTM \cite{hochreiter1997long}, GRU, etc.) have gradually been replaced by 3D-CNN \cite{tran2015learning} and Transformer in video classification. Because their limited ability to model long-range dependencies and lower computational efficiency. Compared to traditional RNN structures, Transformers leverage self-attention mechanisms to effectively capture long-range dependencies. This enables more precise modeling of inter-frame relationships in videos. Moreover, their parallel computing capability significantly improves training efficiency, making them more advantageous for handling large-scale sequential data. 

Motivated by these advantages, the Transformer architecture is adopted for temporal feature processing in temporal feature processing for MsFIN. For the given feature sequences $\widehat{\mathcal{X}}=\left[ {\widehat{x}_{1}},{\widehat{x}_{2}}, \dots,{\widehat{x}_{T}} \right]\in {{\mathbb{R}}^{T \times d}}$, the Causal Temporal Module (CTM) adopts self-attention mechanism, following the same computation process as equations (\ref{equ1}-\ref{equ2}). Additionally, to ensure the causality of accident anticipation, a causal mask is applied after computing the attention scores to prevent future information leakage.

To balance computational cost and predictive performance, a processing method similar to Model3 in \cite{arnab2021vivit} is adopted. Specifically, for the aggregated object-level features, only the temporal evolution of individual objects is considered. $\widehat{\mathcal{O}}$ is reshaped into ${{\mathbb{R}}^{N\times T\times d}}$ and then passed into the CTM. In fact, the relational information between objects has already been modeled during multi-scale feature aggregation layer through the SaM. Therefore, in temporal layer, object interaction are no longer explicitly considered. Instead, the focus is placed on the temporal evolution of object features themselves to extract more stable sequential information.

\begin{figure}[t]
\centering
\includegraphics{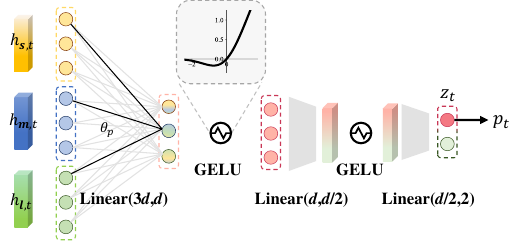}
\caption{MLP for accident probability prediction.}
\label{MLP}
\end{figure}

\subsection{Multi-scale Feature Post Fusion}
Multi-scale post-fusion layer integrates multi-scale frame-level features and object-level features output from the temporal layer to generate the final frame representations. Specifically, multi-scale frame-level features $\widetilde{\mathcal{F}}_{p}$ from the temporal layer serve as query vectors, while the object-level features $\widetilde{\mathcal{O}}$ act as keys and values. These features are fed into three CaMs for interaction, enabling the post-fusion of multi-scale frame-level and object-level features. As illustrated in Fig.~\ref{MLP}, the MLP consists of three linear layers, each followed by a GeLU activation, except for the final output layer. The model outputs the accident probability $p_t$ for each frame.

\begin{equation}
\mathcal{R}=\text{MLP}\left( \text{Concat}({\mathcal{H}_{s}},{\mathcal{H}_{m}},{\mathcal{H}_{l}}) \right)
\end{equation}

\subsection{Loss Function}
In this paper, TAA is formulated as a classification task. For each input driving sequence consisting of $T$ frames at a frame rate of $r$ , positive samples are defined as driving sequences that contain the accident occurrence frame ${{t}_\text{ao}}$ , with video-level classification label of 1. Conversely, negative samples are defined as driving sequences that do not contain ${{t}_\text{ao}}$ , with video-level classification label of 0. During model training, a cross-entropy loss function with an exponential decay factor is employed:

\begin{equation}
\begin{split}
\mathcal{L} = -\sum\limits_{t=1}^{T} \Big[ & \left(1 - y_t\right) \log \left(1 - p_t\right) \\
& + y_t \cdot e^{-\max\left(0, \frac{t_{ao} - t}{r}\right)} \log(p_t) \Big]
\end{split}
\end{equation}

\noindent where ${{p}_{t}}$ is the probability of a positive sample at time step $t$. ${{y}_{t}}$ is the classification label of the driving sequence. For negative samples, the loss function adopts the standard cross-entropy loss, encouraging the model to correctly predict non-accident scene. For positive samples, the exponential decay factor imposes a greater penalty on incorrect predictions that occur closer to the accident frame. This mechanism enhances the model's focus on critical time steps. 

This exponential loss is widely used in other accident anticipation works \cite{liao2024crash,li2024cognitive, chan2017anticipating, karim2022dynamic, zeng2017agent}. However, introducing the exponential decay factor alone cannot fully address the issue of easy-to-classify samples dominating the gradient updates. A large number of easily predicted normal frames cause the loss function to concentrate gradients on these low-difficulty samples. This limits the ability of the model to effectively learn from hard positive examples. To mitigate this issue, focal loss \cite{lin2017focal} is further introduced to focus training on challenging samples. The final loss function for training the MsFIN is defined as:

\begin{equation}
\mathcal{L}_\text{neg}(t) = \alpha \cdot p_t^{\gamma} \cdot (1 - y_t) \log(1 - p_t) 
\end{equation}

\begin{equation}
\mathcal{L}_\text{pos}(t) = (1 - \alpha) \cdot (1 - p_t)^{\gamma} \cdot y_t \cdot e^{- \max\left(0, \frac{t_{ao} - t}{r} \right)} \log(p_t)
\end{equation}

\begin{equation}
\mathcal{L} = - \sum_{t=1}^{T} \left[ \mathcal{L}_\text{neg}(t) + \mathcal{L}_\text{pos}(t) \right]
\end{equation}

\noindent where $\gamma $ serves as a modulation factor to reduce the loss contribution of easily classified samples. And $\alpha $ represents the class weight, adjusting the loss ratio between positive and negative samples.

\section{Experiments}
In this section, the accident anticipation capability of the proposed MsFIN is evaluated through experiments on two publicly available datasets, DAD and DADA. The performance is further compared with that of other models.

\subsection{Datasets}
The DAD dataset \cite{chan2017anticipating} is the first widely used dashcam accident dataset, comprising 620 driving sequences. It is sampled into 1,750 video clips, including 1,120 negative samples and 620 positive samples. Each video clip spans 100 frames at 20 fps, with accidents in positive samples fixed at the 90th frame. Notably, the dataset contains a limited variety of accident types, 58.2\% of accidents involve motorcycles. Real-world accident scene are complex and involve diverse accident types. To assess the robustness of MsFIN more comprehensively, it is further tested on the DADA dataset across multiple accident categories.

The DADA dataset \cite{fang2021dada} consists of 2,000 complete accident videos covering 54 accident types. It contains a total of 658,476 frames, with an average video length of approximately 230 frames at a 30-frame rate. Each video is annotated with comprehensive accident information, including abnormal frames (frames where accident participants first appear), accident occurrence frames, and accident end frames. Additionally, driver attention data is provided for each video frame. The dataset was further enriched with accident descriptions, accident causes, and preventive measures by \cite{li2024cognitive}.

\subsection{Implementation details}
The construction of MsFIN is implemented based on the PyTorch \cite{paszke2019pytorch} framework, with training and testing conducted on an Nvidia RTX 4090 GPU with 24GB of memory.

To ensure a fair comparison of experimental results, the evaluation of the model on DAD dataset directly utilizes the feature data provided by \cite{chan2017anticipating} for training and testing. The feature extraction method is as follows: For each frame, image information is processed using a pre-trained VGG-16 \cite{simonyan2014very} for frame-level feature extraction. Meanwhile, object detection is performed using a fine-tuned Faster R-CNN. The detected objects are then passed through VGG-16 to extract object-level features. Both frame-level and object-level features have a dimensionality of 4,096. To balance computational cost and predictive performance, the number of candidate objects is limited to 19. As a result, the dimensionality of the DAD video sequence feature data is 100×20×4096. Each sequence contains 100 frames (20 fps). And each frame consists of frame-level features along with 19 object-level features, each with a dimensionality of 4,096.

For the DADA dataset, 514 sets of long accident video sequences were selected, including 360 for training and 154 for testing. These sequences cover 39 major accident types. To eliminate the impact of the positive-to-negative sample ratio in the test set on prediction accuracy, each long accident video is divided into two shorter sequences. Each sequence contains 150 frames (30 fps). One serves as a positive sample, and the other as a negative sample. The end frame of the positive sample is randomly selected between $t_\text{ao}$ and $t_\text{ae}$ (accident end frame). The end frame of the negative sample is randomly chosen before $t_\text{ai}$ (abnormal frame). To align with the DAD feature data, frame features and 15 object features were extracted from each video frame using the same method as in \cite{chan2017anticipating}. Consequently, the dimensionality of the DADA video sequence feature data is 150×16×4096. Each sequence contains 150 frames. Each frame consisting of frame-level features and 15 object-level features, each with a dimensionality of 4,096.

Frame-level and object-level features are then reduced to 512 (D) using a linear layer before being fed into the multi-scale feature aggregation layer. In SaM, the short-term window is set to fps/3, while the mid-term window is set to fps. The layers of SaM, CaM and CTM are all configured to two layers. In both experiments, the learning rate for network training is set to 0.0001, the batch size is set to 10. Training is conducted for 60 epochs using the AdamW optimizer.

Existing Vision-TAA methods use standard classification metrics to assess the correctness of accident anticipation. These metrics include Accuracy, Precision (Pre) , Recall (Rec), Average Precision (AP), and the Area Under the ROC Curve (AUC). The Average Precision is calculated as:

\begin{equation}
\text{AP} = \sum_{n=1}^{N} (R_n - R_{n-1}) \cdot P_n
\end{equation}

\noindent whrere $P_n$ and $R_n$ denote the Pre and Rec at the n-th threshold. In addition, Time to Accident (TTA) is used to measure the earliness of accidents anticipate. TTA is defined as the time interval between the frame where the predicted accident risk exceeds a predefined threshold and the actual accident frame. 

\begin{equation}
\text{TTA} = \max \left\{ t_{\text{ao}} - t \mid p_t \geq \tau,\ 1 \leq t \leq t_{\text{ao}} \right\}
\end{equation}

\noindent A variant, mean Time to Accident (mTTA), is typically computed by averaging TTA values across multiple thresholds. It provides a more comprehensive assessment of earliness capability.

\begin{equation}
\text{mTTA} = \frac{1}{|\mathcal{T}|} \sum_{\tau \in \mathcal{T}} \text{TTA}(\tau)
\end{equation}

\noindent where $\mathcal{T}$ is the set of thresholds from 0 to 1.

Four classical metrics are used to evaluate the MsFIN. AP and AP$_{80R}$ for assessing Correctness, as well as mTTA and TTA$_{80R}$ for evaluating Earliness. AP quantifies the overall predictive performance of the model by computing the area under the Precision-Recall curve at different Rec levels. A higher AP value indicates that the model maintains high prediction accuracy across various classification thresholds. AP$_{80R}$ represents the precision corresponding to a Rec of 80\%. This metric evaluates the reliability of predictions under the constraint of a high Rec. It prevents the model from overly prioritizing early predictions at the expense of accuracy. mTTA calculates the mean lead time of correctly predicted positive samples (True Positives) across different thresholds, This metric reflects how early the model can issue warnings before an accident occurs. A higher mTTA value indicates that the model can provide a longer emergency response time. This gives drivers or autonomous agents more time to react, which helps reduce or even prevent accidents. TTA$_{80R}$ represents the TTA value when the Rec reaches 80\%. It balances the reliability and earliness of accident warnings, providing a more comprehensive assessment for the practical deployment value of the model.

\subsection{Experimental Results}

\begin{figure*}[h]
\centering
\includegraphics{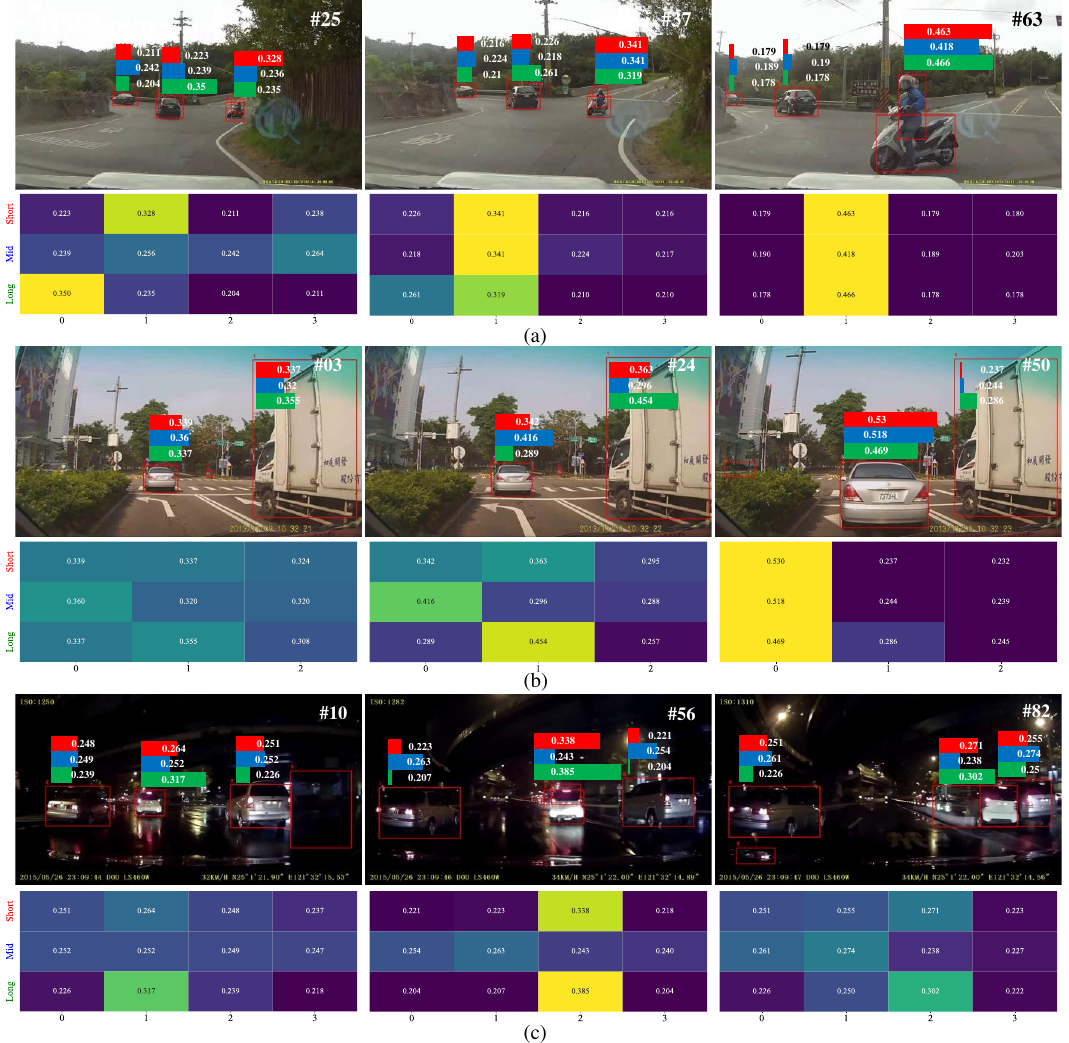}
\caption{The complementarity of Multi-scale features for accident anticipation. (a) represents a sudden event, demonstrating the importance of short-term scale for timely accident anticipation. (b) represents a gradually evolving risk, where mid-term scale plays a key role in capturing the progression. (c) represents a low-visibility condition, highlighting how long-term scale helps to anticipate accidents under such challenging scenes. They demonstrate how multi-scale features complement each other under different accident types.}
\label{Attention_matrix}
\end{figure*}

\begin{figure*}[t]
\centering
\includegraphics{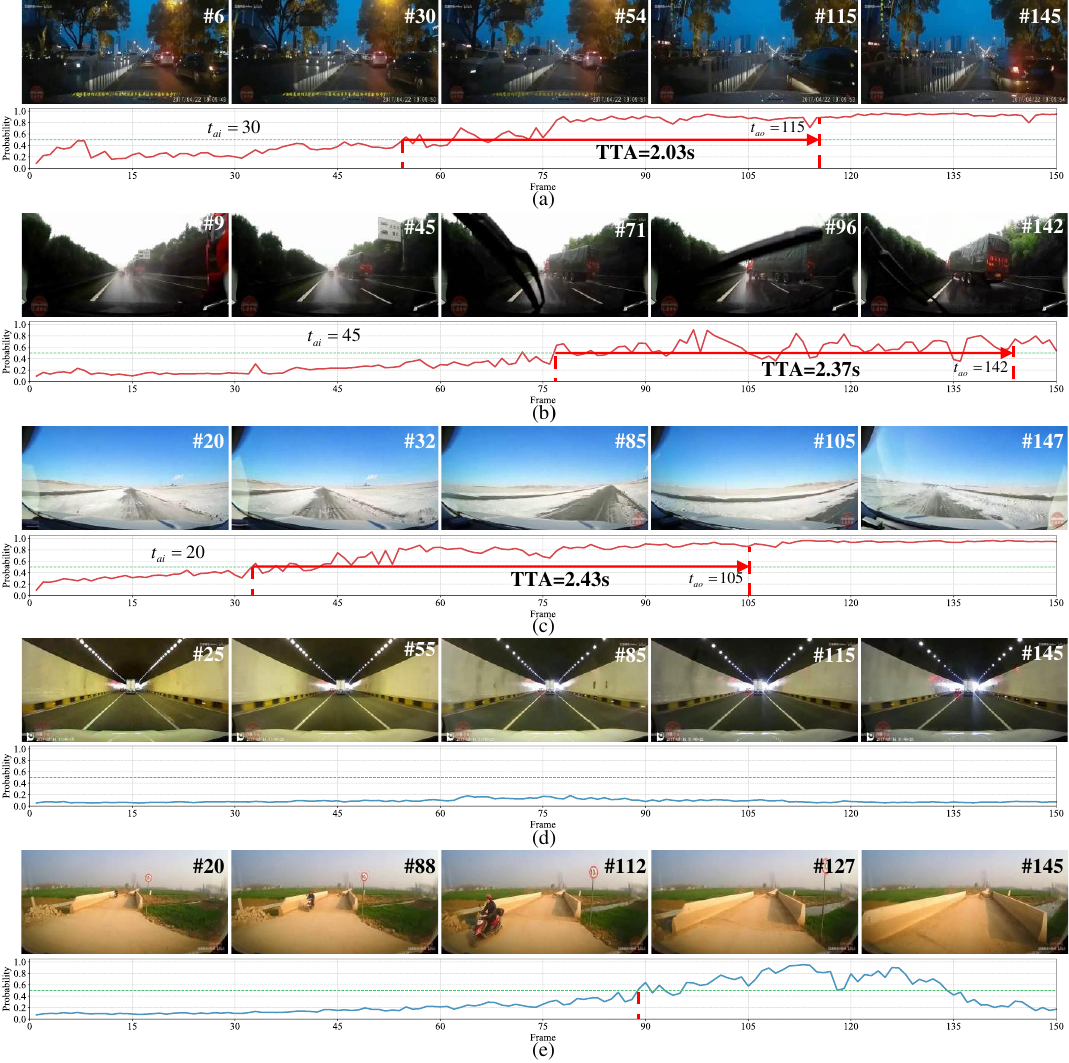}
\caption{Examples of accident anticipation on the DADA dataset. (a), (b), and (c) demonstrate positive sample predictions in different scenes and accident types. (d) illustrates a correctly predicted negative sample, and (e) represents a false negative example. The predicted accident probabilities are displayed at the bottom.}
\label{Probability_curve}
\end{figure*}

\subsubsection{Greater earliness TAA}To better demonstrate the advantages of multi-scale features in TAA, this paper visualizes the attention scores in the multi-scale post-fusion layer, as shown in Fig.~\ref{Attention_matrix}. The image sequence of the driving scene is presented at the top. The attention scores for frame-level features at different temporal scales to object-level features are shown at the bottom. For clarity, bar charts indicating the attention scores from different temporal scales for key participant are displayed above the bounding boxes. Specifically, red, blue, and green bars represent short-term, mid-term, and long-term temporal scales, respectively. 

Compared to single-scale modeling methods, multi-scale features can form complementary relationships when facing different accident types, enabling greater earliness in capturing accident risks. Short-term scales are particularly effective in capturing sudden changes. Fig.~\ref{Attention_matrix} (a) describes cyclist suddenly appears at the intersection and collides with the ego vehicle. The cyclist emerges abruptly at frame 25, and the short-term scales immediately responds to this event with a significant increase in attention score. In contrast, the mid-term and long-term scales do not immediately focus on the cyclist. They shift attention only when the cyclist approaches the ego vehicle and shows noticeable behavioral changes. 

Mid-term scales are more attentive to behavioral changes of participants. In Fig.~\ref{Attention_matrix} (b), a white car ahead slowly reverses and collides with the ego vehicle. Due to the slow motion, neither the short-term nor the long-term scale paid attention to the changes in this vehicle in early stage of the accident. Instead, they focused more on the nearby truck, which poses a more immediate threat. The mid-term scale initiates continuous attention to the reversing vehicle at frame 24. Other scales gradually converging their focus onto this accident vehicle one second later. 

Long-term scales are more effective under low-visibility conditions such as nighttime or rainy weather, where visual cues are subtle. In such scene, the short-term and mid-term scales may fail to capture meaningful changes. The long-term scale preserves the maximum feature change of the participant over a longer time window, compensating for the limitations of visual perception. Fig.~\ref{Attention_matrix} (c) illustrates a white car ahead of the ego vehicle changing lanes to the right and colliding with a vehicle on that side. Because the visual cues of the participants are weak, the short- and mid-term attention is scattered. The long-term scale leverages maximum feature changes from earlier time steps to focus on a specific participant. As shown in the figure, despite at collision frame, other scales show significantly lower attention scores for the involved vehicle compared to the long-term scale.

From the visualization, it is also observed that mid-term attention tends to be more distributed across participants compared to other scales. This is due to the average pooling strategy adopted for the mid-term scale, which enables it to capture broader behavioral trends across multiple participants. Furthermore, in the early stages of accidents, attention from all three scales tends to be more dispersed, while in later stages, attention becomes more focused. This indirectly reflects the process of transitioning from risk exploration to risk localization. In the early stage, the overall scene appears similar to normal traffic, with comparable behavior dynamics among vehicles and no apparent anomalies. Thus, the overall risk level remains low.

\begin{figure*}[t]
\centering
\includegraphics{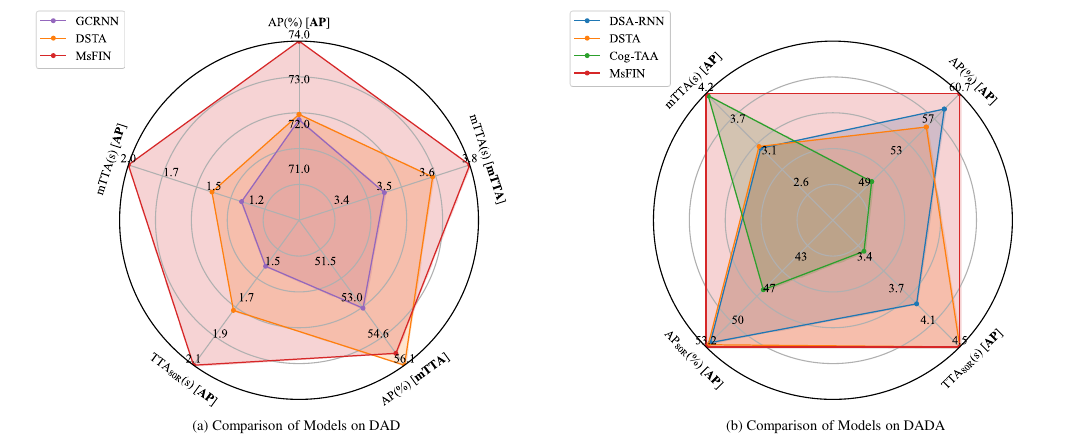}
\caption{Comparison of models on DAD and DADA.}
\label{Quantitative results}
\end{figure*}

\subsubsection{Effectiveness across multiple scenes} Fig.~\ref{Probability_curve} presents five examples of accident anticipation results using the proposed MsFIN. (a), (b), and (c) demonstrate successful positive sample predictions in different scenes and accident types. (d) illustrates a correctly predicted negative sample, and (e) represents a false negative example. For each case, the image sequence of the driving scene is shown at the top. At the bottom, the time series of predicted accident probabilities are displayed. A warning is triggered when the predicted probability exceeds a threshold of 0.5.

Fig.~\ref{Probability_curve} (a), the ego vehicle collides with another vehicle in the night scene of the city road. When the black vehicle begins to change lanes into the ego lane, its lane change behavior appears slow and normal. This leads to a gradual increase in the predicted accident probability, which remains below the threshold. As the vehicle draws closer, the risk becomes more evident. The predicted probability eventually surpasses the threshold at frame 54, resulting in a lead time of 2.03 seconds before the accident occurs.

Fig.~\ref{Probability_curve} (b), during a rainy highway scene, an accident involving another vehicle occurred. A warning is triggered at frame 71, when a forward-facing truck shows signs of tilting, with a TTA of 2.37s. (c) involves the snow-covered road where the ego vehicle skids and loses balance. (d) presents a non-accident scene inside a tunnel. The ego vehicle maintains a safe distance from surrounding vehicles, with the predicted accident probability consistently remaining below 50\%. 

Fig.~\ref{Probability_curve} (e) shows a rural road in sunny weather with no actual accident occurring. As the ego vehicle begins to move forward, an oncoming motorcycle rapidly approaches from the opposite direction. Due to the motorcycle's high speed and close proximity, the model triggers a warning at frame 88. However, since the motorcycle adjusts its trajectory in time to avoid the collision, the predicted risk value subsequently decreases. Notably, although this is a false positive, such conservative prediction behavior is rational in safety-critical traffic scene. This tendency for false alarms essentially reflects the high sensitivity of the system to potential risk.

\subsubsection{Quantitative results} Fig.~\ref{Quantitative results} summarizes the comparison results of MsFIN on the DAD and DADA. Due to the introduction of an exponential decay factor in the positive sample term of the loss function during training, the model is encouraged to anticipate accidents at an earlier stage. However, this design often results in a decrease in accuracy. Conversely, if the model prioritizes prediction accuracy, it tends to trigger predictions closer to the actual accident occurrence, leading to a reduction in mTTA. This trade-off between accuracy and early anticipation suggests an inherent balance in accident prediction. Therefore, on DAD dataset the MsFIN is evaluated against existing single-scale interaction models in terms of both achieving best AP and longer mTTA. The performance of DSTA and GCRNN\cite{bao2020uncertainty} are cited from\cite{karim2022dynamic}. On DAD dataset, DSTA achieves an AP of 72.34\% in the best AP comparison, with a mTTA of 1.50s and a TTAR80 of 1.81s. Compared to DSTA, the proposed MsFIN improves AP by 1.62\%, while extending mTTA and TTAR80 by 0.48s and 0.31s, respectively. Furthermore, in the longer mTTA comparison, the MsFIN achieves an AP of 55.6\% and a mTTA of 3.76s. These results indicate that MsFIN have more balanced and superior performance compared to existing methods. It not only maintains prediction accuracy but also preserves a longer anticipation lead time.

However, early anticipation lacking sufficient accuracy provide limited practical benefits and may even erode driver trust in accident anticipation systems. As a result, when AP is low, comparisons based on longer mTTA become less meaningful. Therefore, the subsequent experiments focuses on models with higher AP. On DADA dataset, the MsFIN achieves a mTTA of 4.25s and a TTA80R of 4.47s, making it the only model with an AP exceeding 60\%. These results indicate that even on the DADA dataset, which features diverse accident types and complex traffic scene, the MsFIN consistently outperforms other models. Notably, although the Cog-TAA \cite{li2024cognitive} surpasses the MsFIN by 0.06s in the mTTA comparison, it suffers a significant accuracy drop of over 10\%. 

\begin{figure}[t]
\centering
\includegraphics{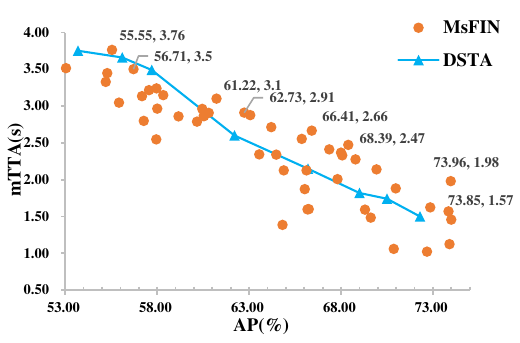}
\caption{The mTTA-AP diagram of the MsFIN.}
\label{mTTA-AP}
\end{figure}

To further compare MsFIN with the DSTA network, the mTTA-AP diagram is presented in Fig.~\ref{mTTA-AP}. As shown in the figure, MsFIN consistently outperforms the DSTA network in both mTTA and AP. Notably, under relatively high AP conditions, the mTTA of MsFIN is approximately 0.5s longer than that of the DSTA network on average. This represents a significant improvement.

\begin{figure}[t]
\centering
\includegraphics{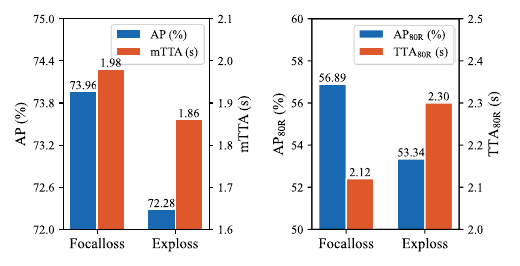}
\caption{Loss function comparison.}
\label{com_loss}
\end{figure}

\begin{table}[t]
  \centering
  \caption{Ablation experiment}
  \label{rem_ablation}
  \setlength{\tabcolsep}{6pt}
  \renewcommand{\arraystretch}{1.2}
  \begin{tabular}{
    >{\centering\arraybackslash}p{1em}
    >{\centering\arraybackslash}p{2.5em}
    >{\centering\arraybackslash}p{1em}
    >{\centering\arraybackslash}p{1em}
    >{\centering\arraybackslash}p{1em}
    >{\centering\arraybackslash}p{2em}
    >{\centering\arraybackslash}p{1em}
    >{\centering\arraybackslash}p{1em}
    >{\centering\arraybackslash}p{2.5em}
    >{\centering\arraybackslash}p{2.5em}
  }
    \toprule
    \multirow{2}{*}{\textbf{Exp}} & \multirow{2}{*}{\textbf{TL}} & \multicolumn{3}{c}{\textbf{MsM}} & \multirow{2}{*}{\textbf{SaM}} & \multicolumn{2}{c}{\textbf{CaM}} & \multirow{2}{*}{\textbf{AP(\%)}} & \multirow{2}{*}{\textbf{TTA(s)}} \\
    & & S & M & L & & pre & pos & & \\
    \midrule
    1 & CTM & \checkmark & \checkmark & \checkmark & \checkmark & \checkmark & \checkmark & \textbf{73.96} & \textbf{1.94} \\
    2 & CTM & -- & \checkmark & \checkmark & \checkmark & \checkmark & \checkmark & 74.55 & 1.74 \\
    3 & CTM & \checkmark & -- & \checkmark & \checkmark & \checkmark & \checkmark & 73.94 & 1.46 \\
    4 & CTM & \checkmark & \checkmark & -- & \checkmark & \checkmark & \checkmark & \textbf{75.02} & 1.63 \\
    5 & CTM & \checkmark & \checkmark & \checkmark & -- & \checkmark & \checkmark & 73.65 & 1.46 \\
    6 & CTM & \checkmark & \checkmark & \checkmark & \checkmark & -- & \checkmark & 71.04 & \textbf{2.01} \\
    7 & CTM & \checkmark & \checkmark & \checkmark & \checkmark & \checkmark & -- & 71.01 & 1.68 \\
    8 & GRU & \checkmark & \checkmark & \checkmark & \checkmark & \checkmark & \checkmark & 73.38 & 1.35 \\
    9 & Mamba & \checkmark & \checkmark & \checkmark & \checkmark & \checkmark & \checkmark & 71.74 & 1.51 \\
    \bottomrule
  \end{tabular}
\end{table}

\subsubsection{Ablation experiment} 
To validate the effectiveness of the main modules, ablation experiments were conducted on the MsFIN framework using the DAD dataset, including both module replacement and module removal. These experiments demonstrate the impact of different module designs and validate the necessity of each component in achieving correctness and earliness in accident anticipation.

Fig.~\ref{com_loss} presents the results of replacing the loss function in MsFIN. Experimental results show that introducing focal loss leads to improvements of 1.68\% in AP, 3.55\% in AP$_{80R}$, and 0.12s in mTTA compared to exponential loss. It is worth noting that even with the exponential loss, MsFIN still outperforms DSTA. 

The experimental results of removing individual modules from the MsFIN architecture are summarized in TABLE~\ref{rem_ablation}. Experiment 1 demonstrates that the complete MsFIN, incorporating all major components, achieves AP of 73.96\% and mTTA of 1.98s. The configurations of each module have been detailed in the previous section.

Experiments 2 to 4 conduct ablation experiments on short-term, mid-term, and long-term scale features respectively. To isolate the effect of individual variables, the ablation of a specific scale feature is carried out solely within the Multi-scale Module. The structure of the subsequent temporal layer remains unaltered during this process. The results reveal that the exclusion of any single-scale feature has limited impact on correctness. Notably, removing the long-term scale features even leads to a 1.06\% improvement in AP. However, the absence of multi-scale features hinders the model’s ability to capture the progression of accidents over time. This leads to a significant reduction in early anticipation capability, with mTTA dropping by more than 0.2 seconds.

Experiments 5 to 7 perform interaction ablation experiments on MsFIN. In Experiment 5, interaction between object-level features are removed. Experiments 6 and 7 respectively eliminate the pre-fusion and post-fusion interaction between object-level and frame-level features. Among these, the post-fusion interaction between object-level and frame-level features has the most significant overall impact on both correctness and earliness. Its removal leads to a 2.95\% drop in AP and a 0.26s reduction in mTTA.

Benefiting from a modular design, MsFIN supports flexible architectural modifications. This modularity allows for the seamless integration of emerging deep learning components, which is particularly advantageous in the rapidly evolving landscape of neural networks. Mamba \cite{gu2023mamba} is a novel State Space Model that differs from traditional RNN and Transformer. It is based on continuous-time state space representations and leverages implicitly parameterized recurrent computation to achieve linear complexity in sequence modeling. To evaluate the effectiveness of different temporal modeling methods, the original CTM is replaced with GRU and Mamba, respectively. Experimental results demonstrate that CTM outperforms both alternatives in terms of accuracy and early anticipation, primarily due to its global interaction capability enabled by the self-attention mechanism.

\section{Conclusion}
This paper proposes the MsFIN, a novel framework for early anticipation of traffic accidents. By comprehensively modeling the feature interactions between multi-scale scene and participants information, the proposed MsFIN significantly enhances both the correctness and earliness of accident anticipation. Experiments on DAD and DADA datasets demonstrate that multi-scale features are complementary in traffic accident anticipation task, effectively improving the earliness of risk prediction. Compared with existing single-scale interaction models, MsFIN achieves state-of-the-art performance with 73.96\% of AP and 1.98s of mTTA. 

Moreover, some false positive samples actually align with the subjective perception of potential risk by the driver within the traffic scene. However, the current evaluation framework is incapable of assessing the latent risk in negative sample sequences. In addition, for some positive sample sequences, there are no observable clues of accident participants in the early stages. This raises the question of whether it is reasonable to issue an early anticipation before any accident-related participants appear. Therefore, how to objectively define the moment when a scene begins to contain risk remains an open question. These two issues limit the practical deployment of accident anticipation systems. Future work will focus on how to objectively reflect the true risk of traffic scene. 

\bibliographystyle{IEEEtran}
\bibliography{MsFIN}

\vspace{-10 mm}
\begin{IEEEbiography}[{\includegraphics[width=1in,height=1.25in,clip,keepaspectratio]{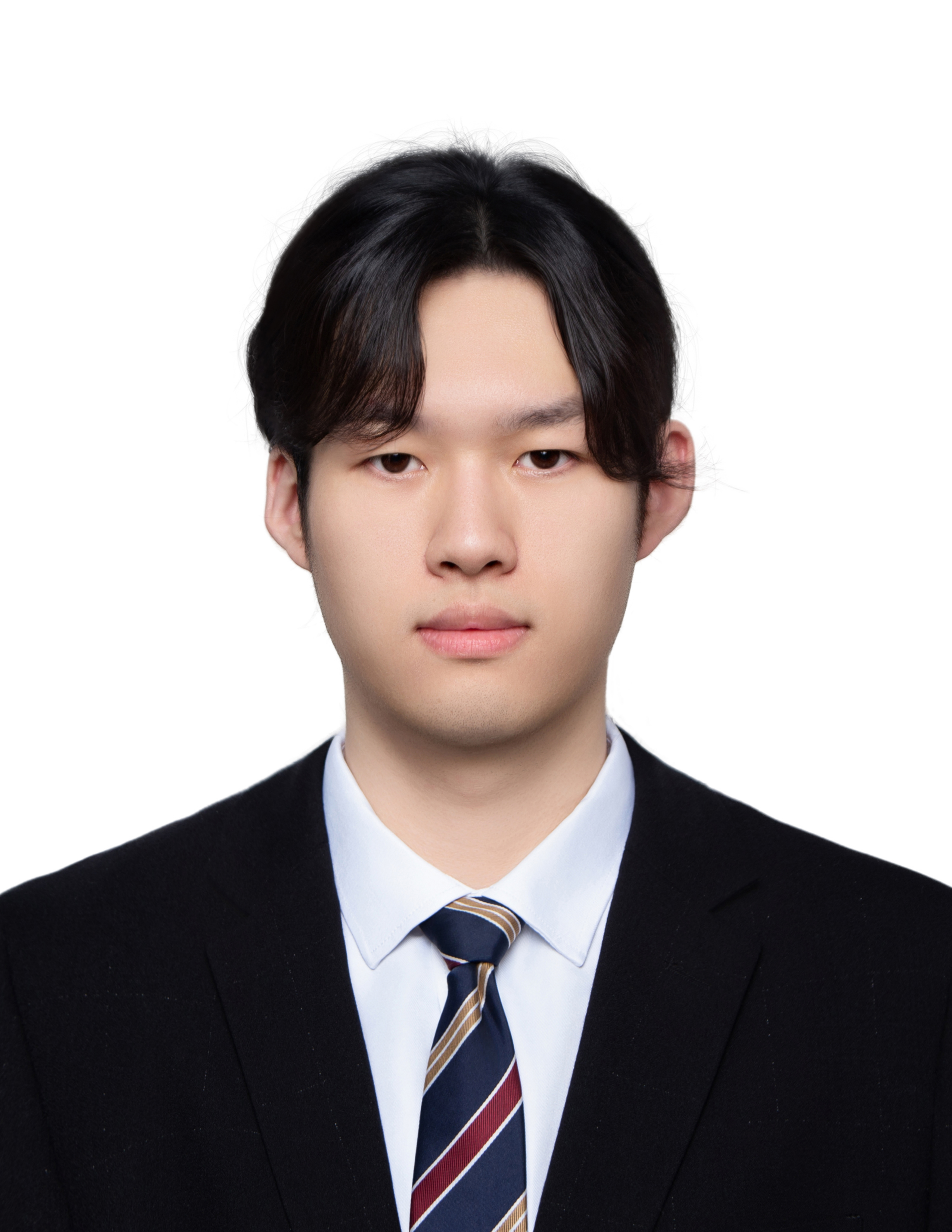}}]{Tongshuai Wu}
received the B.S. degree in vehicle engineering from the Beijing Institute of Technology, Beijing, China, in 2025, where he is currently working toward the M.S. degree with the School of mechanical engineering. His research interests include human-computer interaction, cognitive driving, and path planning of intelligent vehicles.
\end{IEEEbiography}

\vspace{-10 mm}
\begin{IEEEbiography}[{\includegraphics[width=1in,height=1.25in,clip,keepaspectratio]{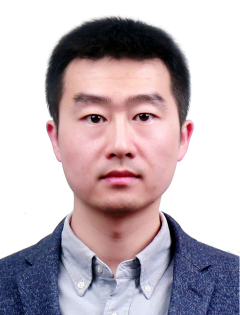}}]{Chao Lu}
received the B.S. degree in transport engineering from the Beijing Institute of Technology (BIT), Beijing, China, in 2009 and the Ph.D. degree in transport studies from the University of Leeds, Leeds, U.K., in 2015. In 2017, he was a Visiting Researcher with the Advanced Vehicle Engineering Centre, Cranfield University, Cranfield, U.K. He is currently an Associate Professor with the School of Mechanical Engineering, BIT. His research interests include intelligent transportation and vehicular systems, driver behavior modeling, reinforcement learning, and transfer learning and its applications.
\end{IEEEbiography}

\vspace{-10 mm}
\begin{IEEEbiography}[{\includegraphics[width=1in,height=1.25in,clip,keepaspectratio]{author/Ze_Song.pdf}}]{Ze Song}
received the B.S. degree in automation from Northwestern Polytechnical University in Xi'an, China, in 2023. He is currently pursuing the M.S. degree at Beijing Institute of Technology. His research interests include interactive behavior modeling and human-like decision-making for intelligent vehicles.
\end{IEEEbiography}

\vspace{-10 mm}
\begin{IEEEbiography}[{\includegraphics[width=1in,height=1.25in,clip,keepaspectratio]{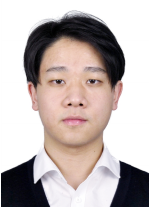}}]{Yunlong Lin}
Yunlong Lin received the B.S. degree in mechanical engineering from the Beijing Institute of Technology (BIT), Beijing, China, in 2022, where he is currently pursuing the Ph.D. degree. His research interests include interactive behavior modeling, continual learning, and decision-making of intelligent vehicles.
\end{IEEEbiography}

\vspace{-10 mm}
\begin{IEEEbiography}[{\includegraphics[width=1in,height=1.25in,clip,keepaspectratio]{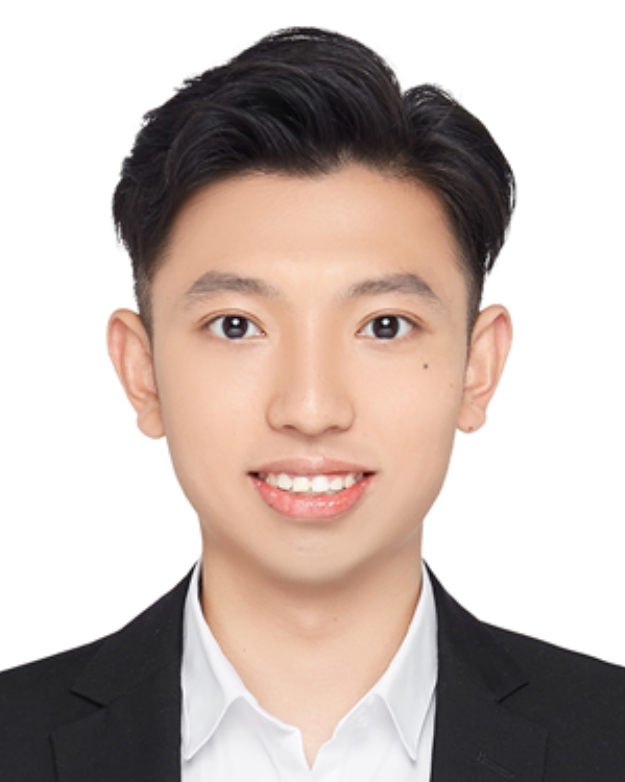}}]{Sizhe Fan}
Sizhe Fan received the B.S. degree in vehicle engineering from Beijing Institute of Technology (BIT), Beijing, China, in 2023, where he is currently pursuing the M.S. degree. His research interests include gaze of drivers, risk analysis and lifelong learning of intelligent vehicles.
\end{IEEEbiography}

\vspace{-10 mm}
\begin{IEEEbiography}[{\includegraphics[width=1in,height=1.25in,clip,keepaspectratio]{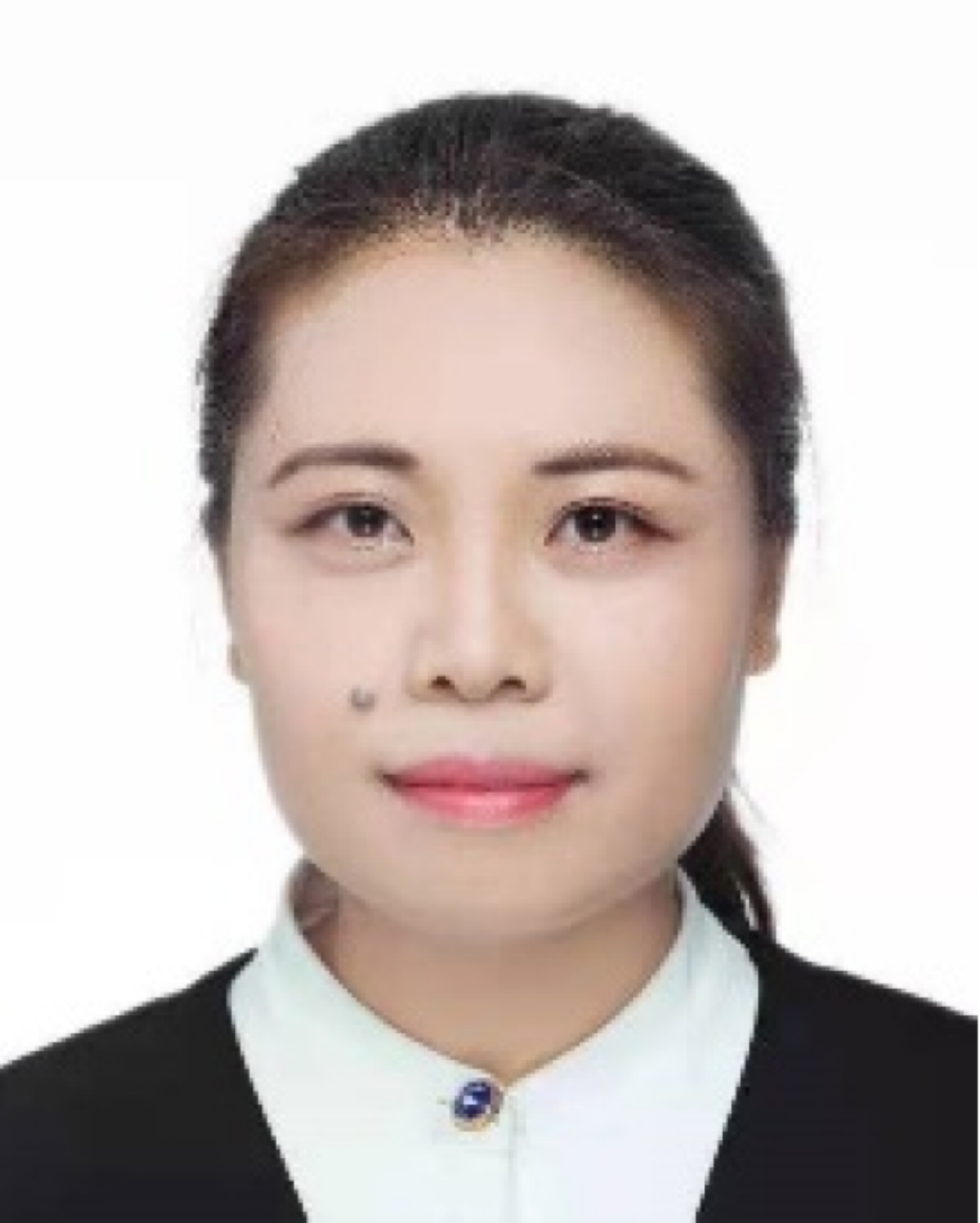}}]{Xuemei Chen}
obtained B.E. in automobile operation engineering from the Shandong University of Technology, Zibo, in 2000 and M.S. degree from the Beijing University of Technology, Beijing, in 2003 and the Ph.D. degree in Beijing Institute of Technology, Beijing, in 2006. She is currently an Associate Professor in the Mechanical Engineering Department, the Beijing Institute of Technology and the Executive Vice president of Advanced Technology Research Institute, Beijing Institute of Technology since 2021. Her research interests include driver behavior model, autonomous vehicle decision-making and machine learning.
\end{IEEEbiography}

\vfill
\end{document}